\documentclass[11pt,a4paper]{article}
\usepackage[hyperref]{naaclhlt2019}
\usepackage{times}
\usepackage{latexsym}
\usepackage{fancyvrb}
\usepackage{upquote}
\usepackage{listings}
\usepackage{url}
\usepackage{graphicx}
\usepackage{todonotes}
\usepackage{xspace}
\usepackage{booktabs}
\usepackage{verbatim}
\usepackage{mygb4e}

\graphicspath{ {./images/} }

\aclfinalcopy 
\def\aclpaperid{1409} 

\lstdefinestyle{PenmanStyle}{
  language=Lisp,
  basicstyle=\ttfamily\footnotesize,
  alsoletter={-+*},
  morekeywords=[2]{
    ARG1-NEQ, carg, -of,
    sf, tense, perf, mood, num, pers, ind
  },
  keywordstyle=[2]{\color{red!50!blue}},
  stringstyle={\itshape\color{green!70!black}},
  linewidth=0.9\linewidth,
  breaklines=true,
  postbreak=\mbox{\textcolor{red}{$\hookrightarrow$}\space},
}

\title{Neural Text Generation from Rich Semantic Representations}

\author{Valerie Hajdik$^1$, Jan Buys$^2$, Michael W. Goodman$^1$ and Emily M.\ Bender$^1$ \\
  $^1$Department of Linguistics, University of Washington \\
  $^2$Paul G. Allen School of Computer Science and Engineering, University of Washington\\
  {\tt \small vhajdik@uw.edu, jbuys@cs.washington.edu,  \{goodmami, ebender\}@uw.edu}
}

\date{}

\begin{document}
\maketitle

\begin{abstract}

We propose neural models to generate high-quality text from structured representations based on Minimal Recursion Semantics (MRS).
MRS is a rich semantic representation that encodes more precise semantic detail than other representations such as Abstract Meaning Representation (AMR).
We show that a sequence-to-sequence model that maps a linearization of Dependency MRS, a graph-based representation of MRS, to English text can achieve a BLEU score of 66.11 when trained on gold data.
The performance can be improved further using a high-precision, broad coverage grammar-based parser to generate a large silver training corpus, achieving a final BLEU score of 77.17 on the full test set, and 83.37 on the subset of test data most closely matching the silver data domain. 
Our results suggest that MRS-based representations are a good choice for applications that need both structured semantics and the ability to produce natural language text as output.

\end{abstract}

\section{Introduction}
Text generation systems often generate their output from an intermediate semantic representation \cite{yao2012semantics,Takase2016neural}.
However many semantic representations are task- or domain-specific \citep{he2003data,wong2007generation}, while rule-based text generation systems often have incomplete coverage \cite{langkilde2002empirical,oepen2007towards}. 

In this work we combine the advantages of Minimal Recursion Semantics \citep[MRS;][]{Copestake2005} with the robustness and fluency of neural sequence-to-sequence models trained on large datasets.
We hypothesize that MRS is particularly well-suited for text generation, as it is explicitly compositional, capturing the contribution to sentence meaning of all parts of the surface form \cite{Bender2015LayersOI}.

\begin{figure}[t]
    \centering
    \begin{tabular}{|c|}
         \hline
          \textbf{DMRS} \\ 
          \includegraphics[scale=0.45]{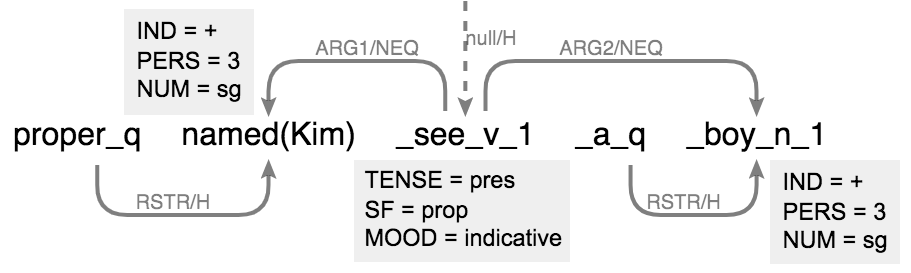} \\
          \hline
          \textbf{AMR}\\
          \includegraphics[scale=0.45]{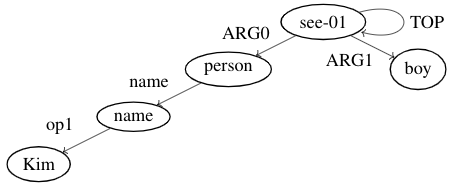} \\
         \hline
    \end{tabular}
    \caption{DMRS and AMR graphs for the sentence \emph{Kim sees a boy}. Because DMRS includes tense and number, and has a node for the determiner, it can distinguish between, e.g.\ \emph{Kim sees a boy} and \emph{Kim saw the boys}, which AMR does not do.}
    \label{table:graphvisualize}
\end{figure}

In contrast, semantic representations such as Abstract Meaning Representation \citep[AMR;][]{banarescu-EtAl:2013:LAW7-ID}  seek to abstract away from the syntax of a sentence as much as possible.
Therefore MRS captures meaning distinctions that AMR fails to represent (see Fig.~\ref{table:graphvisualize}).

Our approach (\S \ref{sec:approach}) uses neural sequence-to-sequence models \cite{Sutskevar2014,AlignAndTranslate2014} to map linearizations of directed acyclic graphs (DAGs) to text, similar to the approach proposed by  \citet{NeuralAMR2017} to generate text from AMR.
We use Dependency MRS \citep[DMRS;][]{Copestake2009}, a graph-based representation in which nodes are MRS predicates (annotated with additional attributes) and edges represent relations between predicates. 
MRS and DMRS are interconvertible and the graph-based representation enables more convenient linearization and manipulation than MRS's variable-based representation \cite{Copestake16Applications}.

Results (\S \ref{sec:results}) show that neural DMRS to English text generation can obtain up to 83.37 BLEU and 32\% exact match, substantially higher than previous work.
In particular, we obtain an 11.6 BLEU improvement through  semi-supervised training using the output of a grammar-based parser, compared to training on gold data only.
In comparison a grammar-based generator obtained 62.05 BLEU, and an approach based on DAG Transducers \cite{DAGTransduction2018} 68.07 BLEU.
Ablation experiments show that node attributes encoding fine-grained morpho-semantic information such as number and tense contribute more than 12 BLEU points.
The highest reported result for AMR generation is 33.8 BLEU \cite{NeuralAMR2017}; on the same dataset our best model obtains 75.8 BLEU. 
While a more detailed meaning representation is harder to produce, our results suggest that MRS could be suitable for text generation applications where precise semantic representations are required.

\section{Approach}
\label{sec:approach}

\subsection{Data}

Our gold training data are parallel MRS and English text corpora, derived from the 1214 release of the Redwoods Treebank \citep{LinGO2002}.\footnote{\url{http://svn.delph-in.net/erg/tags/1214/tsdb/gold}} 
MRS is implemented as the semantic layer of the English Resource Grammar \citep[ERG;][]{Flickinger2000,Flickinger2011}, a broad-coverage, hand-engineered computational grammar of English.
The Redwoods annotation was produced in conjunction with the ERG by parsing each sentence into a forest (discarding unparsable sentences), followed by manual disambiguation \cite{ERG2017}. 

About half of the training data comes from the Wall Street Journal (sections 00-21), while the rest spans a range of domains, including Wikipedia, e-commerce dialogues, tourism brochures, and the Brown corpus. 
The data is split into training, development and test sets with 72,190, 5,288, and 10,201 sentences, respectively.

\subsection{Graph linearization}

\begin{figure}[t]
\footnotesize
    \textbf{PENMAN} 
\begin{lstlisting}[style=PenmanStyle,frame=single]
(10002 / _see_v_1
  :tense PRES
  :sf PROP
  :perf -
  :mood INDICATIVE
  :ARG1-NEQ (10001 / named
    :carg "Kim"
    :pers 3
    :num SG
    :ind +)
  :ARG2-NEQ (10004 / _boy_n_1
    :pers 3
    :num SG
    :ind +
    :RSTR-H-of (10003 / _a_q)))
\end{lstlisting} 

\textbf{Linearization} 
\begin{lstlisting}[style=PenmanStyle,frame=single]
( _see_v_1 mood=INDICATIVE|perf=-|sf=PROP|tense=PRES ARG1-NEQ ( named0 ind=+|num=SG|pers=3 ) ARG2-NEQ ( _boy_n_1 ind=+|num=SG|pers=3 RSTR-H-of ( _a_q ) ) )
\end{lstlisting} 
\caption{The DMRS for the sentence \emph{Kim sees a boy.} in PENMAN format (top) and the linearization used by our model (bottom).}
\label{fig:preprocessing}
\end{figure}

We use PyDelphin\footnote{\url{https://github.com/delph-in/pydelphin}} to convert MRS annotations to DMRS.
In order to apply sequence-to-sequence models to graph-to-text generation, we then linearize the DMRS into PENMAN format (which is also used to represent AMR).
We follow \citet[pp.~82--86]{Goodmami2018} in finding normalized spanning trees through depth-first traversal over the directed acyclic DMRS graphs.\footnote{\url{https://github.com/goodmami/mrs-to-penman}} 
The PENMAN format defines each node once, supports node attributes and edge labels, marks edges whose direction is reversed in the traversal,
and represents edges which are not covered by the spanning tree.

The PENMAN format is processed further to obtain a linearization appropriate as input to sequence-to-sequence models, similar to the approach proposed by \citet{NeuralAMR2017} for AMR linearization (see Fig.~\ref{fig:preprocessing}).
Node variable identifiers are removed, node attributes are concatenated, and named entities are anonymized. 
Predicates that appear only once in the training data are treated as unknowns.
Preprocessing and unknown word handling are described in greater detail in Appendices \ref{app:preprocess} and \ref{app:unks}.

\subsection{Model}

Our neural generator follows the standard encoder-decoder paradigm \cite{AlignAndTranslate2014}.
The encoder is a two-layer bidirectional LSTM.
Predicates  and their attributes are embedded separately; their embeddings are then concatenated \cite{LingFeatures2016}. 
The decoder uses global soft attention for alignment \cite{luong2015effective}, and pointer attention to copy unknown tokens directly to the output  \cite{PointingUnknown2016}. 
The models are trained using Adam \cite{kingma2014adam}. Dropout is applied to non-recurrent connections.
Decoding uses beam search (width 5).
The generator is implemented using OpenNMT-py \cite{OpenNMT2017}.
Hyperparameter details are given in Appendix \ref{app:hyperparams}.
Our code is available online.\footnote{\url{https://github.com/shlurbee/dmrs-text-generation-naacl2019}}

\subsection{Semi-supervised training}

We augment the gold training data with a \emph{silver} dataset generated using ACE,\footnote{ACE version 0.9.25, with the 1214 ERG release, available at \url{http://sweaglesw.org/linguistics/ace}} 
a parser for the ERG, to parse sentences to MRS.
We sample one million sentences from the Gigaword corpus \citep{Gigaword}, restricted to articles published before the year 2000, to match the domain of the Wall Street Journal data. 
The parser failed to parse about 10.3\% of the Gigaword sentences, so these were discarded.
While there are robust MRS parsers \cite{Buys2017, ChenSunWan:2018}, the MRSs they produce are less accurate and not guaranteed to be well-formed.
Our approach thus differs from \citet{NeuralAMR2017}, who used self-training to improve AMR to text generation by iteratively training on larger amounts of data parsed by their neural parser.\footnote{The ACE parser obtained 93.5 Smatch score on parsable sentences \cite{Buys2017}, while the neural AMR parser \cite{NeuralAMR2017} obtained 62.1 Smatch (on a different domain).}

\section{Results}
\label{sec:results}

\begin{table*}[t]
\centering 
\begin{tabular}{llllll} 
 \toprule
 \textbf{Model} & \textbf{BLEU} & \textbf{BLEU} & \textbf{BLEU} & \textbf{Exact} & \textbf{Coverage\%} \\ 
 & \textbf{(All)} & \textbf{(WSJ)} & \textbf{(All overlap)} & \textbf{Match\%} &  \\
 \midrule
 Neural MRS (gold) & 66.11 & 73.12 & 69.27 & 24.09 & 100 \\
 Neural MRS (silver) & 75.43 & 81.76 & 77.13 & 25.82 & 100 \\
 Neural MRS (gold + silver) & 77.17 & 83.37 & 79.15 & 32.07 & 100 \\
 ACE (ERG)  & -- & -- & 62.05 & 15.08 & 78 \\
 DAG transducer \cite{DAGTransduction2018} & -- & 68.07 & -- & -- & 100 \\
 \bottomrule
\end{tabular}
\caption{BLEU and exact-match scores over held-out test set}
\label{table:parser_nn_compare}
\end{table*}

We compare the performance of our neural generator when trained on either gold, silver, or gold and silver data (Table \ref{table:parser_nn_compare}).
Generation quality is primarily evaluated with BLEU \citep{BLEU2002}, using SacreBLEU \citep{BLEU2018}.\footnote{\url{https://github.com/mjpost/sacreBLEU}}
We evaluate the neural models on both the full Redwoods test set (`All') and the WSJ subset. 

The results show that our neural generator obtains very strong performance. 
Semi-supervised training leveraging the ERG parser leads to an 11 BLEU point improvement on Redwoods, comparing to supervised training only.
We found that the best semi-supervised results are obtained by upsampling the gold data so that the gold to silver ratio in training examples is 1:2.
Interestingly, training on silver data performs only slightly worse than training on both gold and silver. 

\subsection{Baselines}

Our baselines are the ERG's grammar-based generator \cite{carroll1999efficient,carroll2005high} and the DAG transducer generator of \citet{DAGTransduction2018}.
To compare our models against the grammar-based generator, implemented in ACE, we need to restrict the evaluation to examples from which ACE is able to generate (`All overlap').\footnote{Despite all test sentences being \emph{parsable} by the ERG, there are gaps in generation coverage, primarily because ACE is unable to generate words outside the grammar's vocabulary.}
In addition to BLEU, we also report exact match accuracy on the overlapping subset. 

Results show that our neural models outperform the grammar-based generator by a large margin.
ACE ranks candidate generations with a discriminative ranker based on structural features over its derivations \cite{velldal-oepen:2006:EMNLP}.
However, it does not use a language model trained on large amounts of text, which would likely improve fluency substantially.

The DAG transducer was trained to generate from Elementary Dependency Structures \citep[EDS;][]{EDS2006}, an MRS-derived representation almost equivalent to DMRS (after edge properties are removed, which Table \ref{table:ablation} shows has an effect of less than 1 BLEU point).
It was evaluated against the same WSJ test set reference generations, but trained using both less gold data (only the WSJ subsection) and less silver data (300K vs 900K sentences).
Our model trained on WSJ gold data performs only slightly worse (65.78 BLEU; see Table \ref{table:domain}) and all our semi-supervised models obtain substantially higher results.

\subsection{Out of domain evaluation}

\begin{table}[t]
    \centering
    \begin{tabular}{lcc}
         \toprule
         & \multicolumn{2}{c}{\textbf{Training Data}} \\
         \textbf{Test domain} & \textbf{WSJ} & \textbf{WSJ + Giga} \\
         \midrule
         WSJ & 65.78 & 83.42 \\
         Brown & 45.00 & 76.99 \\
         Wikipedia & 35.90 & 62.26 \\
         \bottomrule
    \end{tabular}
    \caption{BLEU scores for domain match experiments}
    \label{table:domain}
\end{table}

We evaluate the in- and out-of-domain performance of our approach by training models on either WSJ gold data only, or both WSJ gold data and Gigaword silver data, and evaluating on different domains. 
The results in Table~\ref{table:domain} show that while the generator performs best on test data which matches the training domain (news),  semi-supervised training leads to substantial out-of-domain improvements on the Wikipedia and the Brown corpus portions of the test set.

\subsection{Attribute ablations}

\begin{table}[t]
    \centering
    \begin{tabular}{lc}
         \toprule
         \textbf{Ablation} & \textbf{BLEU} \\
         \midrule
         All attributes & 72.06 \\ 
         No node attributes & 59.37 \\
         No node attr except num, tense & 67.34 \\
         No edge features & 71.27 \\
         \bottomrule
    \end{tabular}
    \caption{Results of semantic feature ablation, model trained with gold data only}
    \label{table:ablation}
\end{table}

To understand which elements of MRS contribute most to our generator's performance, we ablate node (predicate) and edge attributes from both the training and test DMRS graphs (Table~\ref{table:ablation}).
In the training data, number and tense show the most variation among node attributes, and subsequently have the largest effect on the reported BLEU score.
The most common value for number is \textsc{sg}, but 62.36\% of sentences contain a node with \textsc{pl}.
Similarly, 42.41\% of sentences contain a tense value other than \textsc{pres} or \textsc{untensed}. 
Many other attributes are less informative:
Mood has a value other than \textsc{indicative} in only 0.38\% of sentences, and perf is \verb=+= in just 9.74\% of sentences.
Edge features (including H, EQ and NEQ) encode constraints on scopal relationships (see \citealt{Copestake2009}).
Removing them, which makes the DMRS representation close to equivalent to EDS, has only a small impact on performance.

\subsection{Comparison with AMR generation}

We compare our approach to AMR-to-text generation by evaluating our generator on a standard AMR test set (LDC2015E86). 
As we do not have manually verified MRSes available on this test set, we use ACE to parse the reference sentences to silver MRSes.
We then evaluate the outputs that our generator produces from those MRSes.  
About 20\% of the examples could not be parsed by ACE, and are discarded for the MRS evaluation.
We compare our generator to the neural AMR generator of \citet{NeuralAMR2017} for models trained on gold as well as gold plus silver data.\footnote{The AMR and DMRS systems have different gold training data, but the same source of silver data.}  

We evaluate DMRS models both with and without predicate and edge attributes, as these attributes contain information that is absent from AMR.\footnote{Recently, \citet{donatelli2018annotation} proposed adding tense and aspect to AMR, but this annotation is not yet available in a large AMR corpus.} 
The results in Table \ref{table:amr_results} show that our MRS generator performs better than the AMR generator by a large margin, even when the additional MRS attributes are excluded. 
Our system results are reported on the subset for which we obtained MRS parses. AMR results are as given by \citet{NeuralAMR2017} and cover the entire test set.

\begin{table}[t]
\centering
\begin{tabular}{@{}lcc@{}}
\toprule
\textbf{Representation}        & \textbf{Train on} & \textbf{Train on} \\ 
        & \textbf{Gold} & \textbf{Gold+Silver}  \\ 
\midrule
AMR                   & 22.0          & 33.8 \\ 
DMRS - no attributes  & 40.1         & 63.6 \\ 
DMRS - all attributes &  56.9        & 75.8 \\ 
\bottomrule
\end{tabular}
\caption{BLEU scores for evaluating AMR and DMRS generators on an AMR test set} 
\label{table:amr_results}
\end{table}

\subsection{Error analysis} 
\label{ssec:error-analysis}

We sampled 99 items for error analysis from the dev set, 33 each from among sentences with sentence-level BLEU scores of 80-89, 60-69, and 40-49.\footnote{Items with BLEU scores lower than 40 tend to be very short and primarily involve formatting differences.} 
We identified all differences between these strings and the reference strings
and then labeled each difference with a fine-grained error type.\footnote{This
was done by a single annotator only. The labels were generated bottom up, with new labels added as needed in the course of annotation.}
We classified the differences into 238 errors, distributed across five levels of severity (Table~\ref{tab:err}).

\begin{table}[t]
\footnotesize
\begin{tabular}{lrrrr}
\toprule
\textbf{Type} & 
\multicolumn{1}{l}{\textbf{B80-89}} &
\multicolumn{1}{l}{\textbf{B60-69}} &
\multicolumn{1}{l}{\textbf{B40-49}} &
\multicolumn{1}{l}{\textbf{All}}\\ 
\midrule
Unproblematic & 56.4 & 39.55 & 48.8 & 47.1\\
Slightly \\
problematic & 18.0 & 9.2 & 3.3 & 7.6\\
Moderately\\
problematic & 12.8 & 25.0 & 18.7 & 19.8\\
Ungrammatical & 5.1 & 7.9 & 8.1 & 7.6\\
Other serious\\
error & 7.7 & 18.4 & 21.1 & 18.1\\ 
\midrule
Number of errors & 39 & 76 & 123 & 238\\
Errors per item & 1.18 & 2.30 & 3.73 & 7.21\\
\bottomrule
\end{tabular}
\caption{Percentage of errors of each type, across 99 sampled items, grouped by BLEU score}
\label{tab:err}
\end{table}

Almost half of the differences (47.1\%) were unproblematic, including spelling variants, meaning-preserving punctuation variation and grammatical alternations (such as optional \emph{that} or auxiliary contraction as in (\ref{ex:1})).  
The slightly problematic category includes close synonyms (e.g.\ \emph{sometime} v.\ \emph{someday}), spelled out number names where Arabic numerals are preferred, and differences in formatting. 
The next more serious category (moderately problematic) includes meaning-changing differences in punctuation, tense or aspect, and minor grammatical errors such as swapping \emph{who} and \emph{which} in relative clauses or \emph{a} v.\ \emph{an}.

\begin{exe}\small
\item\label{ex:1}
\begin{xlist}
\item I think I would like a Sony. [sys.]
\item I think I'd like a Sony. [ref.]
\end{xlist}
\end{exe}

Finally, among the most serious errors, we find cases where the generator provided ungrammatical output or grammatical output not conveying the correct semantics. The former include spurious additional tokens, ungrammatical word orders, and ungrammatical inflection. 
Serious errors that nonetheless resulted in grammatical strings include meaning-changing dropped or swapped tokens (as in (\ref{ex:2})), spurious additional tokens, and word order changes that alter the semantics or pragmatics of the string.

\begin{exe}\small
\item\label{ex:2}
\begin{xlist}
        \item {For such cases, machine learning techniques emulate human linguistics and learn from training examples to predict future events.} [sys.]
        \item {For such cases, machine learning techniques emulate human cognition and learn from training examples to predict future events.} [ref.]
\end{xlist}
\end{exe}

In summary, we find that the BLEU scores underestimate the quality of system
outputs, due to unproblematic differences (N=112) and differences, like formatting markup (N=6), not reflected in the
input semantic representations.
Among the 108 moderate to serious differences, about a third (35)
involve punctuation, suggesting that meaning signalled by punctuation could be better reflected in the semantic representations. 
About half (52) involve added, dropped, or swapped tokens, showing room for improvement in the generator's ability to learn appropriate connections between semantic predicates and surface forms. 
The remainder (21) involve inflection, grammatical alternations (such as \emph{who}/\emph{which}) and word order constraints, showing room for improvement in mimicking grammatical processes.

\section{Conclusion}

We have shown that neural sequence-to-sequence models can be used to generate high quality natural language text from Minimal Recursion Semantics representations, in contrast to both existing MRS-based generators and neural generators based on other broad-coverage semantic representations.
Furthermore, we have demonstrated that a large hand-crafted grammar can be leveraged to produce large training sets, which improves performance of neural generators substantially. 
Therefore we argue that the ability to generate high quality text from MRS makes it a good choice of representation for text generation applications that require semantic structure.
For future work, we are interested in applying graph-to-sequence neural networks \cite{beck2018graph,song2018graphtoseq} to MRS-to-text generation. 

\section*{Acknowledgements}
Thanks to Yannis Konstas for sharing preliminary results on DMRS generation, and
Swabha Swayamdipta for discussions.
This research was supported in part by NSF (IIS-1524371) and Samsung AI Research. 

\bibliographystyle{acl_natbib}
\bibliography{text_gen_dmrs}

\appendix

\section{Preprocessing Details}
\label{app:preprocess}

We preprocess the text by removing special formatting characters such as HTML tags, converting all double quotations to the same format (i.e., " instead of \verb_''_) and removing double brackets `[[' and `]]' that represent hyperlinks in Wikipedia data. 
These formatting characters are somewhat arbitrary and represent content (markup) that is orthogonal to the meaning of the text, so would be difficult for a model (or a human) to predict from a semantic representation for the text alone (such as MRS). 
After preprocessing and normalization, sentences in the gold and silver training data identical to sentences in the test set were removed.

Named entities (as annotated in MRS) are anonymized according to the conventions described in \citet{NeuralAMR2017} prior to training.
The purpose of anonymization is to reduce the sparsity of tokens such as dates and named entities that, while having a different surface form almost every time they appear in the text, should be treated similarly by the model. These tokens are replaced, both in the DMRS graph and in the raw sentence text, with numbered placeholders like \emph{named0}, \emph{named1}, and \emph{month}.

Sentences are tokenized using the NLTK's implementation \citep{NLTK:2009} of Moses-style tokenization \citep{Koehn:et-al:2007}.
During post-processing, generated sentences are de-anonymized (by replacing the placeholder with the original surface form) and de-tokenized, reversing the NLTK tokenization.

\section{Unknown Word Handling}
\label{app:unks}

We use a combination of two methods to deal with unknown words: anonymizing words to a special \emph{UNK0} token and using the pointer attention to copy the token with highest attention value directly from the source to the target~\cite{PointingUnknown2016}. 
There are two types of unknown words in our setup: words that are not recognized by the model because they do not appear in the training data and words that may or may not appear in the training data that are not recognized by the ERG because they are not part of its lexicon. 
Inspection shows that words missing from the ERG lexicon are often misspellings, uncommon proper names, or domain-specific terminology.

For words that are not part of the ERG lexicon, if the word appears only once in the training data, we replace it with \emph{UNK0}. 
Otherwise, if the word appears multiple times in the training set, we include its surface form (instead of a predicate) as the node token to enable the model to learn from it. 
In early experiments, we also tried either anonymizing all unknowns or including all unknowns (relying on the pointer copy mechanism), and found that anonymizing singletons gave best performance while also keeping the vocabulary size reasonable.

\section{Model Details}
\label{app:hyperparams}

The encoder and decoder each have two 800-dimension Bidirectional LSTM layers. 
The main input embedding layer has 500 dimensions and the size of the attribute embedding layer varies between 22 and 25 depending on the number of attribute combinations in the training data. 
We train using negative log likelihood loss, optimizing using Adam with initial learning rate = 0.001, $\beta_1$ = 0.9 and $\beta_2$ = 0.999. 
For regularization, we apply dropout of 0.3 to the non-recurrent connections of the LSTM layers and after the attention layer. 
The number of layers, per-layer hidden state size, and dropout rates were all chosen empirically to minimize validation perplexity.

\end{document}